\documentclass[10pt,twocolumn,letterpaper]{article}
\pdfoutput=1
\usepackage{cvpr}
\usepackage{times}
\usepackage{epsfig}
\usepackage{graphicx}
\usepackage{amsmath}
\usepackage{amssymb}
\usepackage{comment}

\usepackage[pagebackref=true,breaklinks=true,letterpaper=true,colorlinks,bookmarks=false]{hyperref}

\newcommand{\OURS}{ScanComplete}
\newcommand{\CNN}{3D CNN}

\newcommand{\IGNORE}[1]{{}}

\def \path{\bp C}

\cvprfinalcopy 

\def\cvprPaperID{2835} 

\ifcvprfinal\fi
\begin{document}

\title{\OURS: Large-Scale Scene Completion and \\ Semantic Segmentation for 3D Scans}

\author{
	\hspace{-0.5cm} \hfil Angela Dai$^{1,3,5}$~~~~Daniel Ritchie$^{2}$~~~~Martin Bokeloh$^{3}$~~~~Scott Reed$^{4}$~~~~J\"urgen Sturm$^{3}$~~~~Matthias Nie{\ss}ner$^{5}$ \vspace{0.1cm} \\
	\hspace{-0.5cm} \hfil$^{1}$Stanford University~~~~~$^{2}$Brown University~~~~~$^{3}$Google~~~~~$^{4}$DeepMind~~~~~$^{5}$Technical University of Munich \vspace{0.1cm} \\
}

\twocolumn[{%
	\renewcommand\twocolumn[1][]{#1}%
	\maketitle
	{\begin{center}
		\vspace{-0.8cm}
		\includegraphics[width=\linewidth]{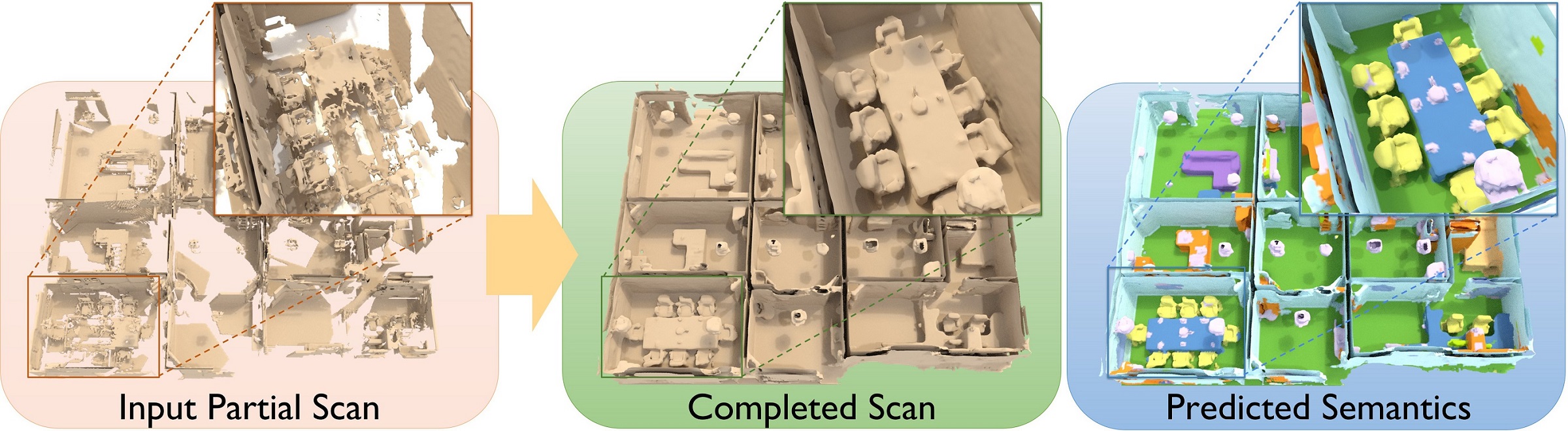}
		\vspace{-0.8cm}
	\end{center}
	}
    3D scans of indoor environments suffer from sensor occlusions, leaving 3D reconstructions with highly incomplete 3D geometry (left). We propose a novel data-driven approach based on fully-convolutional neural networks that transforms incomplete signed distance functions (SDFs) into complete meshes at unprecedented spatial extents (middle). In addition to scene completion, our approach infers semantic class labels even for previously missing geometry (right).
    Our approach outperforms existing approaches both in terms of completion and semantic labeling accuracy by a significant margin.
	\label{fig:teaser}
	\vspace{0.5cm}
}]

\maketitle

\begin{abstract}
\vspace{-2em}
We introduce \OURS{}, a novel data-driven approach for taking an incomplete 3D scan of a scene as input and predicting a complete 3D model along with per-voxel semantic labels.
The key contribution of our method is its ability to handle large scenes with varying spatial extent, managing the cubic growth in data size as scene size increases.
To this end, we devise a fully-convolutional generative 3D CNN model whose filter kernels are invariant to the overall scene size.
The model can be trained on scene subvolumes but deployed on arbitrarily large scenes at test time.
In addition, we propose a coarse-to-fine inference strategy in order to produce high-resolution output while also leveraging large input context sizes.
In an extensive series of experiments, we carefully evaluate different model design choices, considering both deterministic and probabilistic models for completion and semantic inference.
Our results show that we outperform other methods not only in the size of the environments handled and processing efficiency, but also with regard to completion quality and semantic segmentation performance by a significant margin.
\end{abstract}

\section{Introduction}
\label{sec:intro}

With the wide availability of commodity RGB-D sensors such as Microsoft Kinect, Intel RealSense, and Google Tango, 3D reconstruction of indoor spaces has gained momentum \cite{newcombe2011kinectfusion,izadi2011kinectfusion,niessner2013hashing,whelan2015elasticfusion,dai2017bundlefusion}.
3D reconstructions can help create content for graphics applications, and virtual and augmented reality applications rely on obtaining high-quality 3D models from the surrounding environments.
Although significant progress has been made in tracking accuracy and efficient data structures for scanning large spaces, the resulting reconstructed 3D model quality remains unsatisfactory.

One fundamental limitation in quality is that, in general, one can only obtain partial and incomplete reconstructions of a given scene, as scans suffer from occlusions and the physical limitations of range sensors. 
In practice, even with careful scanning by human experts, it is virtually impossible to scan a room without holes in the reconstruction. 
Holes are both aesthetically unpleasing and can lead to severe problems in downstream processing, such as 3D printing or scene editing, as it is unclear whether certain areas of the scan represent free space or occupied space.
Traditional approaches, such as Laplacian hole filling \cite{sorkine2004least,nealen2006laplacian,zhao2007robust} or Poisson Surface reconstruction \cite{kazhdan2006poisson,kazhdan2013screened} can fill small holes.
However, completing high-level scene geometry, such as missing walls or chair legs, is much more challenging.

One promising direction towards solving this problem is to use machine learning for completion.
Very recently, deep learning approaches for 3D completion and other generative tasks involving a single object or depth frame have shown promising results \cite{riegler2017octnetfusion,tatarchenko2017octree,hane2017hierarchical,han2017complete,dai2017complete}.
However, generative modeling and structured output prediction in 3D remains challenging.
When represented with volumetric grids, data size grows cubically as the size of the space increases, which severely limits resolution.
Indoor scenes are particularly challenging, as they are not only large but can also be irregularly shaped with varying spatial extents.

In this paper, we propose a novel approach, \OURS{}, that operates on large 3D environments without restrictions on spatial extent.
We leverage fully-convolutional neural networks that can be trained on smaller subvolumes but applied to arbitrarily-sized scene environments at test time.
This ability allows efficient processing of 3D scans of very large indoor scenes: we show examples with bounds of up to $1480 \times 1230 \times 64$ voxels ($\approx 70 \times 60 \times 3$m).
We specifically focus on the tasks of scene completion and semantic inference:
for a given partial input scan, we infer missing geometry and predict semantic labels on a per-voxel basis.
To obtain high-quality output, the model must use a sufficiently high resolution to predict fine-scale detail.
However, it must also consider a sufficiently large context to recognize large structures and maintain global consistency.
To reconcile these competing concerns, we propose a coarse-to-fine strategy in which the model predicts a multi-resolution hierarchy of outputs.
The first hierarchy level predicts scene geometry and semantics at low resolution but large spatial context.
Following levels use a smaller spatial context but higher resolution, and take the output of the previous hierarchy level as input in order to leverage global context.

In our evaluations, we show scene completion and semantic labeling at unprecedented spatial extents.
In addition, we demonstrate that it is possible to train our model on synthetic data and transfer it to completion of real RGB-D scans taken from commodity scanning devices.
Our results outperform existing completion methods and obtain significantly higher accuracy for semantic voxel labeling.

In summary, our contributions are \vspace{-0.2cm}
\begin{itemize}\setlength{\itemsep}{-1.0mm}
	\item 3D fully-convolutional completion networks for processing 3D scenes with arbitrary spatial extents.
	\item A coarse-to-fine completion strategy which captures both local detail and global structure.
	\item Scene completion and semantic labeling, both of outperforming existing methods by significant margins.
\end{itemize}

\section{Related Work}
\label{sec:relatedWork}

\paragraph{3D Shape and Scene Completion}
Completing 3D shapes has a long history in geometry processing and is often applied as a post-process to raw, captured 3D data.
Traditional methods typically focus on filling small holes by fitting local surface primitives such planes or quadrics, or by using a continuous energy minimization \cite{sorkine2004least,nealen2006laplacian,zhao2007robust}.
Many surface reconstruction methods that take point cloud inputs can be seen as such an approach, as they aim to fit a surface and treat the observations as data points in the optimization process; e.g., Poisson Surface Reconstruction~\cite{kazhdan2006poisson,kazhdan2013screened}.

Other shape completion methods have been developed, including approaches that leverage symmetries in meshes or point clouds \cite{thrun2005shape,mitra2006partial,pauly2008discovering,sipiran2014approximate,speciale2016symmetry} or part-based structural priors derived from a database~\cite{sung2015data}.
One can also `complete' shapes by replacing scanned geometry with aligned CAD models retrieved from a database
\cite{nan2012search,shao2012interactive,kim2012acquiring,li2015database,shi2016data}.
Such approaches assume exact database matches for objects in the 3D scans, though this assumption can be relaxed by allowing modification of the retrieved models, e.g., by non-rigid registration such that they better fit the scan \cite{pauly2005example,rock2015completing}.

To generalize to entirely new shapes, data-driven structured prediction methods show promising results.
One of the first such methods is Voxlets~\cite{firman2016structured}, which uses a random decision forest to predict unknown voxel neighborhoods. 

\paragraph{Deep Learning in 3D}
With the recent popularity of deep learning methods, several approaches for shape generation and completion have been proposed.
3D ShapeNets~\cite{shapenet2015} learns a 3D convolutional deep belief network from a shape database.
This network can generate and complete shapes, and also repair broken meshes \cite{nguyen2016field}.

Several other works have followed, using 3D convolutional neural networks (CNNs) for object classification~\cite{maturana2015voxnet,qi2016volumetric} or completion \cite{dai2017complete,han2017complete}.
To more efficiently represent and process 3D volumes, hierarchical 3D CNNs have been proposed \cite{riegler2017OctNet,wang2017cnn}.
The same hierarchical strategy can be also used for generative approaches which output higher-resolution 3D models \cite{riegler2017octnetfusion,tatarchenko2017octree,hane2017hierarchical,han2017complete}.
One can also increase the spatial extent of a 3D CNN with dilated convolutions \cite{yu2015multi}.
This approach has recently been used for predicting missing voxels and semantic inference \cite{song2017ssc}.
However, these methods operate on a fixed-sized volume whose extent is determined at training time.
Hence, they focus on processing either a single object or a single depth frame.
In our work, we address this limitation with our new approach, which is invariant to differing spatial extent between train and test, thus allowing processing of large scenes at test time while maintaining a high voxel resolution.

\section{Method Overview}
\label{sec:overview}

\begin{figure*}[!htb] 
	\vspace{-0.8cm}
	\centering
	\includegraphics[width=0.95\linewidth]{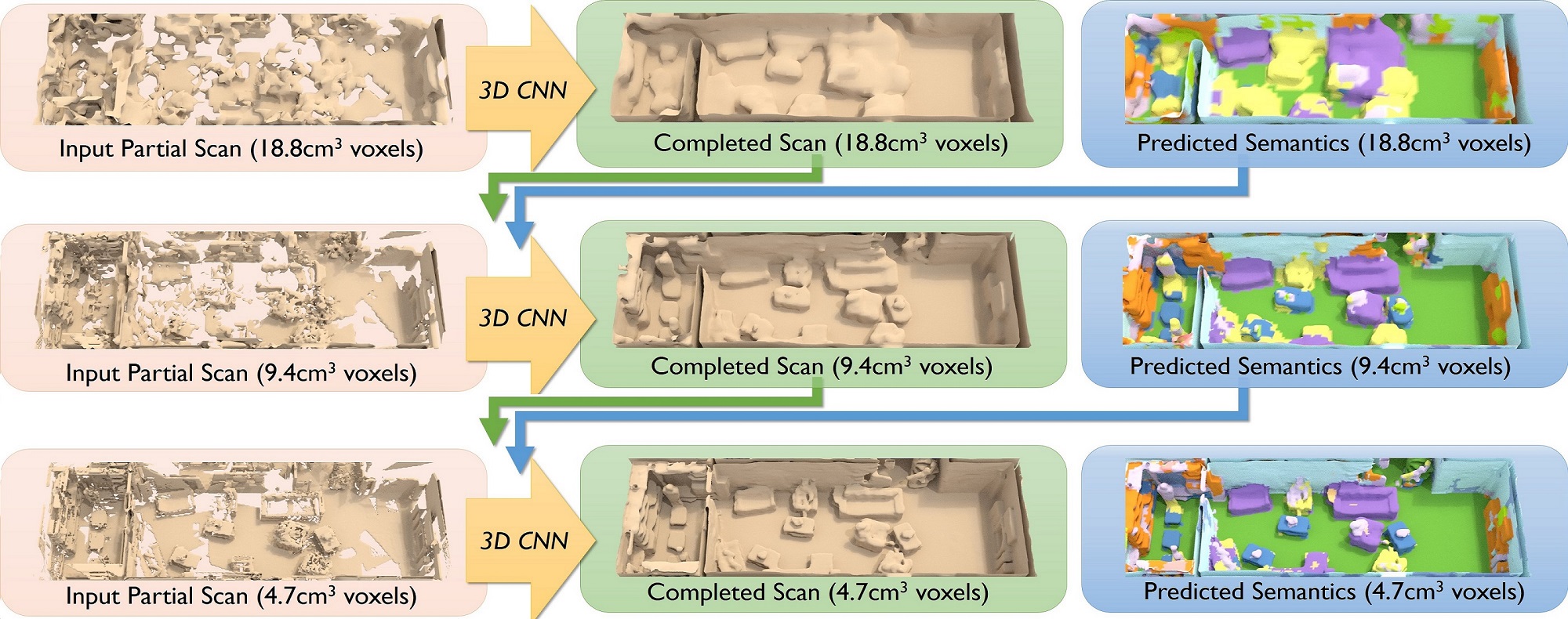}
	\caption{Overview of our method: we propose a hierarchical coarse-to-fine approach, where each level takes a partial 3D scan as input, and predicts a completed scan as well as per-voxel semantic labels at the respective level's voxel resolution using our autoregressive \CNN{} architecture (see Fig.~\ref{fig:network}).
	The next hierarchy level takes as input the output of the previous levels (both completion and semantics), and is then able to refine the results. 
	This process allows leveraging a large spatial context while operating on a high local voxel resolution. In the final result, we see both global completion, as well as local surface detail and high-resolution semantic labels.
}
	\label{fig:overview}
	\vspace{-0.2cm}
\end{figure*}

Our \OURS{} method takes as input a partial 3D scan, represented by a truncated signed distance field (TSDF) stored in a volumetric grid.
The TSDF is generated from depth frames following the volumetric fusion approach of Curless and Levoy \cite{curless1996volumetric}, which has been widely adopted by modern RGB-D scanning methods \cite{newcombe2011kinectfusion,izadi2011kinectfusion,niessner2013hashing,kahler2015very,dai2017bundlefusion}.
We feed this partial TSDF into our new volumetric neural network, which outputs a truncated, unsigned distance field (TDF).
At train time, we provide the network with a target TDF, which is generated from a complete ground-truth mesh.
The network is trained to output a TDF which is as similar as possible to this target complete TDF.

Our network uses a fully-convolutional architecture with three-dimensional filter banks.
Its key property is its invariance to input spatial extent, which is particularly critical for completing large 3D scenes whose sizes can vary significantly.
That is, we can train the network using random spatial crops sampled from training scenes, and then test on different spatial extents at test time.

The memory requirements of a volumetric grid grow cubically with spatial extent, which limits manageable resolutions.
Small voxel sizes capture local detail but lack spatial context; large voxel sizes provide large spatial context but lack local detail.
To get the best of both worlds while maintaining high resolution, we use a coarse-to-fine hierarchical strategy.
Our network first predicts the output at a low resolution in order to leverage more global information from the input.
Subsequent hierarchy levels operate at a higher resolution and smaller context size.
They condition on the previous level's output in addition to the current-level incomplete TSDF.
We use three hierarchy levels, with a large context of several meters ($\sim 6\textrm{m}^3$) at the coarsest level, up to a fine-scale voxel resolution of $\sim 5\textrm{cm}^3$; see Fig.~\ref{fig:overview}.

\begin{figure}[t] 
	\includegraphics[width=\linewidth]{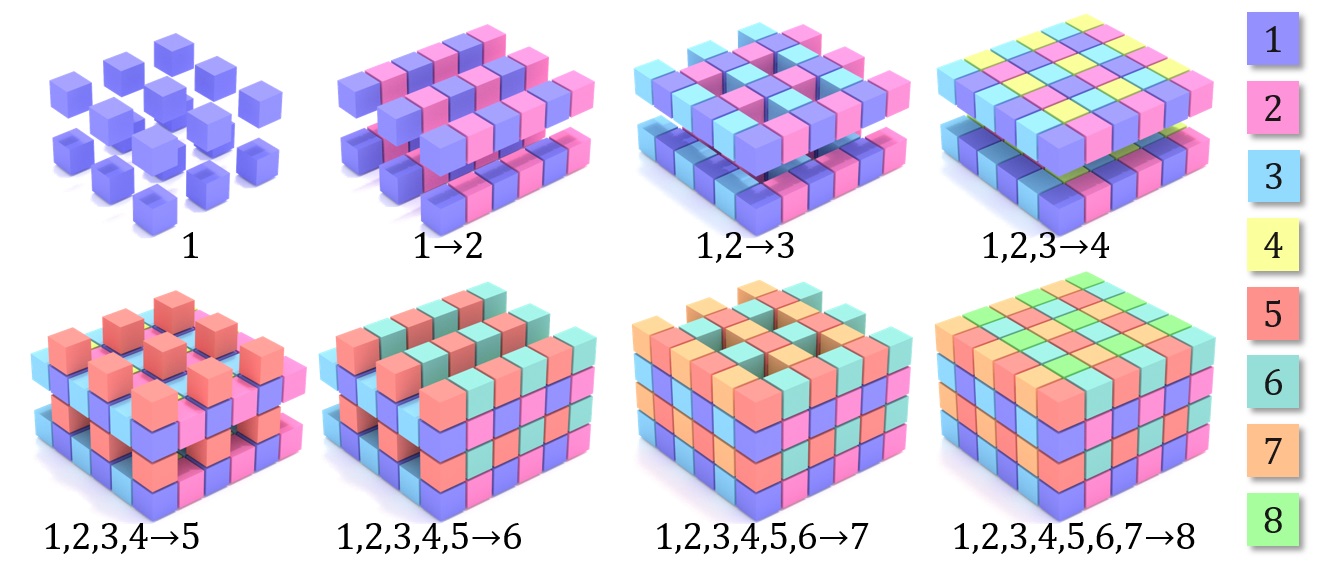}
	\caption{Our model divides volumetric space into eight interleaved voxel groups, such that voxels from the same group do not neighbor each other. It then predicts the contents of these voxel groups autoregressively, predicting voxel group $i$ conditioned on the predictions for groups $1 \ldots i-1$. This approach is based on prior work in autoregressive image modeling~\cite{reedParallel2017}.}
	\label{fig:voxel_groups}
	\vspace{-0.5cm}
\end{figure}

Our network uses an autoregressive architecture based on that of Reed et al.~\cite{reedParallel2017}.
We divide the volumetric space of a given hierarchy level into a set of eight voxel groups, such that voxels from the same group do not neighbor each other; see Fig.~\ref{fig:voxel_groups}.
The network predicts all voxels in group one, followed by all voxels in group two, and so on.
The prediction for each group is conditioned on the predictions for the groups that precede it.
Thus, we use eight separate networks, one for each voxel group; see Fig.~\ref{fig:voxel_groups}.

We also explore multiple options for the training loss function which penalizes differences between the network output and the ground truth target TDF.
As one option, we use a deterministic $\ell_1$-distance, which forces the network to focus on a single mode.
This setup is ideal when partial scans contain enough context to allow for a single explanation of the missing geometry.
As another option, we use a probabilistic model formulated as a classification problem, i.e., TDF values are discretized into bins and their probabilities are weighted based on the magnitude of the TDF value.
This setup may be better suited for very sparse inputs, as the predictions can be multi-modal.

In addition to predicting complete geometry, the model jointly predicts semantic labels on a per-voxel basis.
The semantic label prediction also leverages the fully-convolution autoregressive architecture as well as the coarse-to-fine prediction strategy to obtain an accurate semantic segmentation of the scene.
In our results, we demonstrate how completion greatly helps semantic inference.

\begin{figure*}[htb!]
\vspace{-0.6cm}
	\centering
	\includegraphics[width=0.92\textwidth]{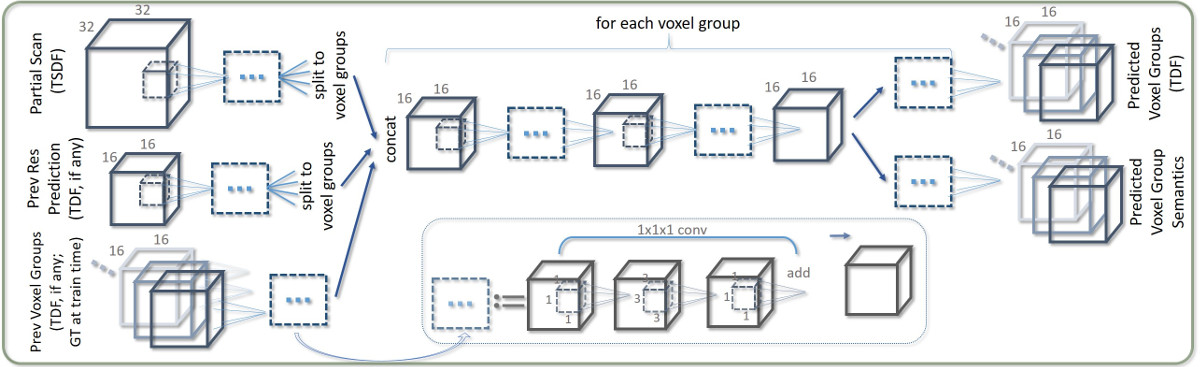}
	\caption{Our \OURS{} network architecture for a single hierarchy level. We take as input a TSDF partial scan, and autoregressively predict both the completed geometry and semantic segmentation.  Our network trains for all eight voxel groups in parallel, as we use ground truth for previous voxel groups at train time. In addition to input from the current hierarchy level, the network takes the predictions (TDF and semantics) from the previous level (i.e., next coarser resolution as input), if available; cf.~Fig.~\ref{fig:overview}. 
	}
	\label{fig:network}
\vspace{-0.4cm}
\end{figure*}

\section{Data Generation}
\label{sec:datagen}

To train our \OURS{} CNN architecture, we prepare training pairs of partial TSDF scans and their complete TDF counterparts.
We generate training examples from SUNCG \cite{song2017ssc}, using 5359 train scenes and 155 test scenes from the train-test split from prior work~\cite{song2017ssc}.
As our network requires only depth input, we virtually scan depth data by generating scanning trajectories mimicking real-world scanning paths.
To do this, we extract trajectory statistics from the ScanNet dataset \cite{dai2017scannet} and compute the mean and variance of camera heights above the ground as well as the camera angle between the look and world-up vectors.
For each room in a SUNCG scene, we then sample from this distribution 
to select a camera height and angle.

Within each $1.5\textrm{m}^3$ region in a room, we select one camera to add to the training scanning trajectory.
We choose the camera $c$ whose resulting depth image $D(c)$ is most similar to depth images from ScanNet.
To quantify this similarity, we first compute the histogram of depth of values $H(D(c))$ for all cameras in ScanNet, and then compute the average histogram, $\bar{H}$.
We then compute the Earth Mover's Distance between histograms for all cameras in ScanNet and $\bar{H}$, i.e., $\text{EMD}(H(D(c)), \bar{H})$ for all cameras $c$ in ScanNet.
We take the mean $\mu_{\text{EMD}}$ and variance $\sigma^2_{\text{EMD}}$ of these distance values.
This gives us a Gaussian distribution over distances to the average depth histogram that we expect to see in real scanning trajectories.
For each candidate camera $c$, we compute its probability under this distribution, i.e., $\mathcal{N}(\text{EMD}(H(D(c)), \bar{H}), \mu_{\text{EMD}}, \sigma_{\text{EMD}})$.
We take a linear combination of this term with the percentage of pixels in $D(c)$ which cover scene objects (i.e., not floor, ceiling, or wall), reflecting the assumption that people tend to focus scans on interesting objects rather than pointing a depth sensor directly at the ground or a wall.
The highest-scoring camera $c^*$ under this combined objective is added to the training scanning trajectory.
This way, we encourage a realistic scanning trajectory, which we use for rendering virtual views from the SUNCG scenes.

For rendered views, we store per-pixel depth in meters.
We then volumetrically fuse \cite{curless1996volumetric} the data into a dense regular grid, where each voxel stores a truncated signed distance value.
We set the truncation to $3 \times$ the voxel size, and we store TSDF values in voxel-distance metrics.
We repeat this process independently for three hierarchy levels, with voxel sizes of $4.7\textrm{cm}^3$, $9.4\textrm{cm}^3$, and $18.8\textrm{cm}^3$.

We generate target TDFs for training using complete meshes from SUNCG.
To do this, we employ the level set generation toolkit by Batty~\cite{battysdf}.
For each voxel, we store a truncated distance value (no sign; truncation of $3 \times$ voxel size), as well as a semantic label of the closest object to the voxel center.
As with TSDFs, TDF values are stored in voxel-distance metrics, and we repeat this ground truth data generation for each of the three hierarchy levels.

For training, we uniformly sample subvolumes at 3m intervals out of each of the train scenes.
We keep all subvolumes containing any non-structural object voxels (e.g., tables, chairs), and randomly discard subvolumes that contain only structural voxels (i.e., wall/ceiling/floor) with 90\% probability.
This results in a total of $225,414$ training subvolumes.
We use voxel grid resolutions of $[ 32 \times 16 \times 32 ]$, $[ 32 \times 32 \times 32 ]$, and $[ 32 \times 64 \times 32 ]$ for each level, resulting in spatial extents of $[ 6\textrm{m} \times 3\textrm{m} \times 6\textrm{m} ]$, $[ 3\textrm{m}^3 ]$, $[ 1.5\textrm{m} \times 3\textrm{m} \times 1.5\textrm{m} ]$, respectively.
For testing, we test on entire scenes.
Both the input partial TSDF and complete target TDF are stored as uniform grids spanning the full extent of the scene, which varies across the test set.
Our fully-convolutional architecture allows training and testing on different sizes and supports varying training spatial extents.

Note that the sign of the input TSDF encodes known and unknown space according to camera visibility, i.e., voxels with a negative value lie behind an observed surface and are thus unknown.
In contrast, we use an unsigned distance field (TDF) for the ground truth target volume, since all voxels are known in the ground truth.
One could argue that the target distance field should use a sign to represent space inside objects.
However, this is infeasible in practice, since the synthetic 3D models from which the ground truth distance fields are generated are rarely watertight.
The use of implicit functions (TSDF and TDF) rather than a discrete occupancy grid allows for better gradients in the training process; this is demonstrated by a variety of experiments on different types of grid representations in prior work \cite{dai2017complete}.
\section{\OURS{} Network Architecture}
\label{sec:arch}

Our \OURS{} network architecture for a single hierarchy level is shown in Fig.~\ref{fig:network}.
It is a fully-convolutional architecture operating directly in 3D, which makes it invariant to different training and testing input data sizes.

At each hierarchy level, the network takes the input partial scan as input (encoded as an TSDF in a volumetric grid) as well as the previous low-resolution TDF prediction (if not the base level) and any previous voxel group TDF predictions.
Each of the input volumes is processed with a series of 3D convolutions with $1\times 1\times 1$ convolution shortcuts.
They are then all concatenated feature-wise and further processed with 3D convolutions with shortcuts.
At the end, the network splits into two paths, one outputting the geometric completion, and the other outputting semantic segmentation, which are measured with an $\ell_1$ loss and voxel-wise softmax cross entropy, respectively.
An overview of the architectures between hierarchy levels is shown in Fig.~\ref{fig:overview}.

\subsection{Training}
\label{sec:training}

To train our networks, we use the training data generated from the SUNCG dataset as described in Sec.~\ref{sec:datagen}.

At train time, we feed ground truth volumes as the previous voxel group inputs to the network.
For the previous hierarchy level input, however, we feed in volumes predicted by the previous hierarchy level network.
Initially, we trained on ground-truth volumes here, but found that this tended to produce highly over-smoothed final output volumes. 
We hypothesize that the network learned to rely heavily on sharp details in the ground truth volumes that are sometimes not present in the predicted volumes, as the network predictions cannot perfectly recover such details and tend to introduce some smoothing.
By using previous hierarchy level predicted volumes as input instead, the network must learn to use the current-level partial input scan to resolve details, relying on the previous level input only for more global, lower-frequency information (such as how to fill in large holes in walls and floors).
The one downside to this approach is that the networks for each hierarchy level can no longer be trained in parallel.
They must be trained sequentially, as the networks for each hierarchy level depend on output predictions from the trained networks at the previous level.
Ideally, we would train all hierarchy levels in a single, end-to-end procedure.
However, current GPU memory limitations make this intractable.

Since we train our model on synthetic data, we introduce height jittering for training samples to counter overfitting, jittering every training sample in height by a (uniform) random jitter in the range $[0, 0.1875]$m.
Since our training data is skewed towards walls and floors, we apply re-weighting in the semantic loss, using a 1:10 ratio for structural classes (e.g. wall/floor/ceiling) versus all other object classes.

For our final model, we train all networks on a NVIDIA GTX 1080, using the Adam optimizer~\cite{kingma2014adam} with learning rate $0.001$ (decayed to $0.0001$)
We train one network for each of the eight voxel groups at each of the three hierarchy levels, for a total of 24 trained networks.
Note that the eight networks within each hierarchy level are trained in parallel, with a total training time for the full hierarchy of $\sim 3$ days.

\begin{table*}[htb!]
	\vspace{-0.5cm}
	\begin{center}
		\small
		\begin{tabular}{| c | c | c | c || c | c | c | c |}
			\hline
			Hierarchy & Probabilistic/ & Autoregressive & Input &  $\ell_1$-Err & $\ell_1$-Err & $\ell_1$-Err & $\ell_1$-Err \\ 
			Levels & Deterministic &  & Size & (entire) & (pred. surf.) & (target surf.) & (unk. space) \\ \hline
			1 & prob. (\#quant=256) & non-autoreg. & 32 & 0.248 & 0.311 & 0.969 & 0.324 \\ \hline
			1 & prob. (\#quant=256) & autoreg. & 16 & 0.226 & {\bf 0.243} & 0.921 & 0.290\\ \hline
			1 & prob. (\#quant=256) & autoreg. & 32 & 0.218 & 0.269 & 0.860 & 0.283\\ \hline
			1 & prob. (\#quant=32) & autoreg. & 32 & 0.208 & 0.252 & 0.839 & 0.271 \\ \hline
			1 & prob. (\#quant=16) & autoreg. & 32 & 0.212 & 0.325 & 0.818 & 0.272 \\ \hline
			1 & prob. (\#quant=8) & autoreg. & 32 & 0.226 & 0.408 & 0.832 & 0.284 \\ \hline
			1 & det. & non-autoreg. & 32 & 0.248 & 0.532 & 0.717 & 0.330 \\ \hline
			1 & det. & autoreg. & 16 & 0.217 & 0.349 & 0.808 & 0.282\\ \hline
			1 & det. & autoreg. & 32 & 0.204 & 0.284 & 0.780 & 0.266 \\ \hline
			\hline
			3 (gt train) & prob. (\#quant=32) & autoreg. & 32 & 0.336 & 0.840 & 0.902 & 0.359 \\ \hline
			3 (pred. train) & prob. (\#quant=32) & autoreg. & 32 & 0.202 & 0.405 & 0.673 & 0.251 \\ \hline
			3 (gt train) & det. & autoreg. & 32 & 0.303 & 0.730 & 0.791 & 0.318 \\ \hline
			3 (pred. train) & det. & autoreg. & 32 & {\bf 0.182} & 0.419 & {\bf 0.534} & {\bf 0.225} \\ \hline
		\end{tabular}
		\caption{Quantitative scene completion results for different variants of our completion-only model evaluated on synthetic SUNCG ground truth data. We measure the $\ell_1$ error against the ground truth distance field (in voxel space, up to truncation distance of $3$ voxels). Using an autoregressive model with a three-level hierarchy and large input context size gives the best performance.}
		\label{tab:quant_completion_archVariants}		
	\end{center}
\end{table*}

\begin{table*}[htb!]
\begin{center}
	\small
	\begin{tabular}{| c || c | c | c | c |}
		\hline
		Method &  $\ell_1$-Err & $\ell_1$-Err & $\ell_1$-Err & $\ell_1$-Err \\ 
		 & (entire) & (pred. surf.) & (target surf.) & (unk. space) \\ \hline
        Poisson Surface Reconstruction~\cite{kazhdan2006poisson,kazhdan2013screened} & 0.531 & 1.178 & 1.695 & 0.512 \\ \hline
        SSCNet~\cite{song2017ssc} & 0.536 & 1.106 & 0.931 & 0.527 \\ \hline
        3D-EPN (unet)~\cite{dai2017complete} & 0.245 & 0.467 & 0.650 & 0.302 \\ \hline
        {\bf Ours} (completion $+$ semantics) & 0.202 & 0.462 & 0.569 & 0.248 \\ \hline
        {\bf Ours} (completion only) & {\bf 0.182} & {\bf 0.419} & {\bf 0.534} & {\bf 0.225} \\ \hline
	\end{tabular}
	\caption{Quantitative scene completion results for different methods on synthetic SUNCG data.  We measure the $\ell_1$ error against the ground truth distance field in voxel space, up to truncation distance of $3$ voxels (i.e., 1 voxel corresponds to $4.7\textrm{cm}^3$). Our method outperforms others in reconstruction error.}
	\label{tab:quant_completion_priorWorkComparison}		
\end{center}
\vspace{-0.6cm}
\end{table*}

\section{Results and Evaluation}
\label{sec:results}

\paragraph{Completion Evaluation on SUNCG}
We first evaluate different architecture variants for geometric scene completion in Tab.~\ref{tab:quant_completion_archVariants}.
We test on 155 SUNCG test scenes, varying the following architectural design choices:
\begin{itemize}\setlength{\itemsep}{-1.0mm}
	\item{\textbf{Hierarchy Levels:} our three-level hierarchy (\emph{3}) vs. a single $4.7$cm-only level (\emph{1}). For the three-level hierarchy, we compare training on ground truth volumes (\emph{gt train}) vs. predicted volumes (\emph{pred. train}) from the previous hierarchy level.}
    \item{\textbf{Probabilistic/Deterministic:} a probabilistic model (\emph{prob.}) that outputs per-voxel a discrete distribution over some number of quantized distance value bins (\emph{\#quant}) vs. a deterministic model that outputs a single distance value per voxel (\emph{det.})}.
    \item{\textbf{Autoregressive:} our autoregressive model that predicts eight interleaved voxel groups in sequence (\emph{autoreg.}) vs. a non-autoregressive variant that predicts all voxels independently (\emph{non-autoreg.}).}
    \item{\textbf{Input Size:} the width and depth of the input context at train time, using either \emph{16} or \emph{32} voxels}
\end{itemize}
We measure completion quality using $\ell_1$ distances with respect to the entire target volume (\emph{entire}), predicted surface (\emph{pred. surf.}), target surface (\emph{target surf.}), and unknown space (\emph{unk. space}).
Using only a single hierarchy level, an autoregressive model improves upon a non-autoregressive model, and reducing the number of quantization bins from $256$ to $32$ improves completion (further reduction reduces the discrete distribution's ability to approximate a continuous distance field).
Note that the increase in \emph{pred. surf.} error from the hierarchy is tied to the ability to predict more unknown surface, as seen by the decrease in \emph{unk. space} error.
Moreover, for our scene completion task, a deterministic model performs better than a probabilistic one, as intuitively we aim to capture a single output mode---the physical reality behind the captured 3D scan.
An autoregressive, deterministic, full hierarchy with the largest spatial context provides the highest accuracy.

\begin{figure*}[htb!]
	\vspace{-0.5cm}
	\centering
	\includegraphics[width=0.9\textwidth]{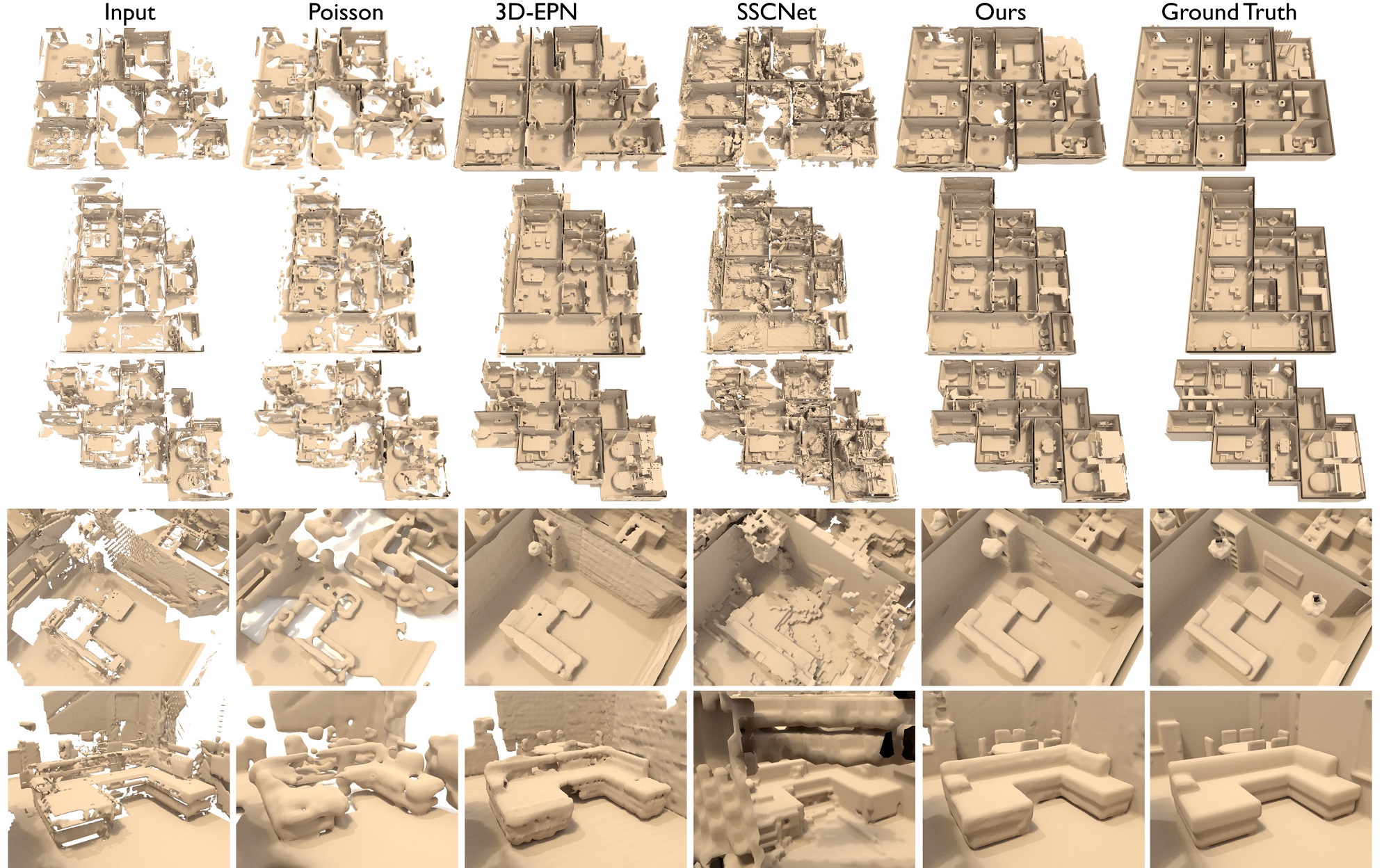}
	\caption{Completion results on synthetic SUNCG scenes; left to right: input, Poisson Surface Reconstruction \cite{kazhdan2013screened}, 3D-EPN \cite{dai2017complete}, SSCNet \cite{song2017ssc}, Ours, ground truth.}
	\label{fig:results_completion_synth}
\end{figure*}

\begin{table*}[htb!]
	\begin{center}
		\small
		\begin{tabular}{| c | c | c | c | c | c | c | c | c | c | c | c || c |}
			\hline
			& bed & ceil. & chair & floor & furn. & obj. & sofa & table & tv & wall & wind. & avg\\ \hline
			(vis) ScanNet~\cite{dai2017scannet} & 44.8 & 90.1 & 32.5 & 75.2 & 41.3 & 25.4 & 51.3 & 42.4 & 9.1 & 60.5 & 4.5 & 43.4\\ \hline
			(vis) SSCNet~\cite{song2017ssc} & 67.4 & 95.8 & 41.6 & 90.2 & 42.5 & 40.7 & 50.8 & 58.4 & 20.2 & 59.3 & {\bf 49.7} & 56.1\\ \hline
			(vis) Ours [sem-only, no hier] & 63.6 & 92.9 & 41.2 & 58.0 & 27.2 & 19.6 & 55.5 & 49.0 & 9.0 & 58.3 & 5.1 & 43.6\\ \hline
			(is) Ours [sem-only] &  {\bf 82.9} & 96.1 & 48.2 & 67.5 & {\bf 64.5} & 40.8 & {\bf 80.6} & 61.7 & 14.8 & 69.1 & 13.7 & 58.2 \\ \hline
			(vis) Ours [no hier] & 70.3 & 97.6 & 58.9 & 63.0 & 46.6 & 34.1 & 74.5 & {\bf 66.5} & {\bf 40.9} & {\bf 86.5} & 43.1 & 62.0\\ \hline
			(vis) {\bf Ours} & 80.1 & {\bf 97.8} & {\bf 63.4} & {\bf 94.3} & 59.8 & {\bf 51.2} & 77.6 & 65.4 & 32.4 & 84.1 & 48.3 & {\bf 68.6}\\ \hline
			\hline
			(int) SSCNet~\cite{song2017ssc} & 65.6 & 81.2 & 48.2 & 76.4 & 49.5 & 49.8 & 61.1 & 57.4 & 14.4 & 74.0 & 36.6 & 55.8\\ \hline
			(int) Ours [no hier] & 68.6 & 96.9 & 55.4 & 71.6 & 43.5 & 36.3 & 75.4 & {\bf 68.2} & {\bf 33.0} & {\bf 88.4} & 33.1 & 60.9 \\ \hline
			(int) {\bf Ours} & {\bf 82.3} & {\bf 97.1} & {\bf 60.0} & {\bf 93.2} & {\bf 58.0} &  {\bf 51.6} & {\bf 80.6} & 66.1 & 26.8 & 86.9 & {\bf 37.3} & {\bf 67.3} \\ \hline
		\end{tabular}
		\caption{Semantic labeling accuracy on SUNCG scenes. We measure per-voxel class accuracies for both the voxels originally visible in the input partial scan (\emph{vis}) as well as the voxels in the intersection of our predictions, SSCNet, and ground truth (\emph{int}). Note that we show significant improvement over a semantic-only model that does not perform completion (\emph{sem-only}) as well as the current state-of-the-art.}
		\label{tab:quant_semantic}	
		\vspace{-0.5cm}	
	\end{center}
\end{table*}

We also compare our method to alternative scene completion methods in Tab.~\ref{tab:quant_completion_priorWorkComparison}.
As a baseline, we compare to Poisson Surface Reconstruction~\cite{kazhdan2006poisson,kazhdan2013screened}.
We also compare to 3D-EPN, which was designed for completing single objects, as opposed to scenes~\cite{dai2017complete}.
Additionally, we compare to SSCNet, which completes the subvolume of a scene viewed by a single depth frame~\cite{song2017ssc}.
For this last comparison, in order to complete the entire scene, we fuse the predictions from all cameras of a test scene into one volume, then evaluate $\ell_1$ errors over this entire volume.
Our method achieves lower reconstruction error than all the other methods.
Note that while jointly predicting semantics along with completion does not improve on completion, Tab.~\ref{tab:quant_semantic} shows that it significantly improves semantic segmentation performance.

We show a qualitative comparison of our completion against state-of-the-art methods in Fig.~\ref{fig:results_completion_synth}.
For these results, we use the best performing architecture according to Tab.~\ref{tab:quant_completion_archVariants}.
We can run our method on arbitrarily large scenes as test input, thus predicting missing geometry in large areas even when input scans are highly partial, and producing more complete results as well as more accurate local detail.
Note that our method is $\mathcal{O}(1)$ at test time in terms of forward passes; we run more efficiently than previous methods which operate on fixed-size subvolumes and must iteratively make predictions on subvolumes of a scene, typically $\mathcal{O}(wd)$ for a $w\times h\times d$ scene.

\begin{figure*}[htb!]
	\centering
	\includegraphics[width=0.8\textwidth]{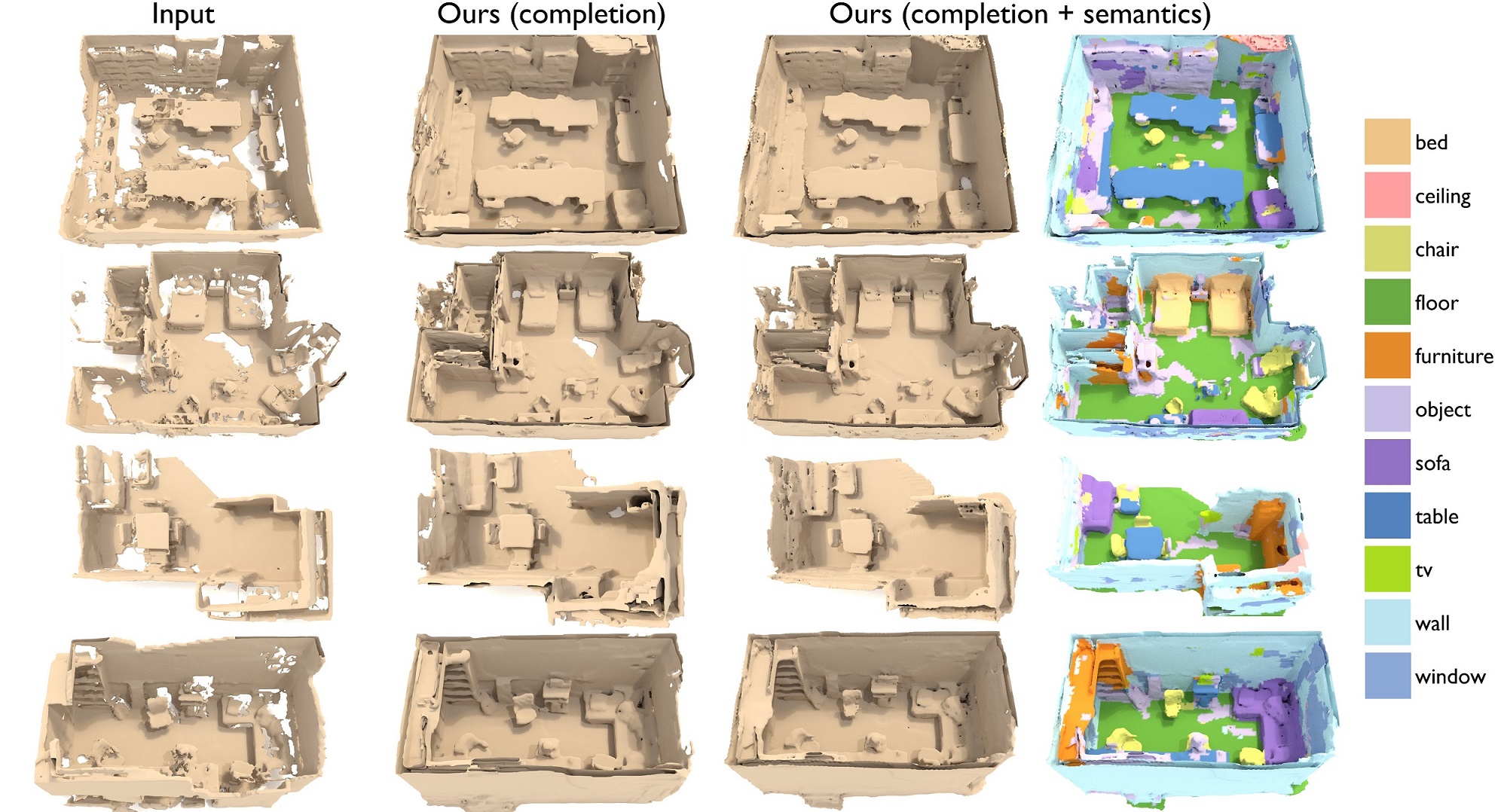}
	\caption{Completion results on real-world scans from ScanNet~\cite{dai2017scannet}. Despite being trained only on synthetic data, our model is also  able to complete many missing regions of real-world data.}
	\label{fig:results_real}
\end{figure*}

\vspace{-0.3cm}
\paragraph{Completion Results on ScanNet (real data)}
We also show qualitative completion results on real-world scans in Fig.~\ref{fig:results_real}.
We run our model on scans from the publicly-available RGB-D ScanNet dataset~\cite{dai2017scannet}, which has data captured with an Occiptal Structure Sensor, similar to a Microsoft Kinect or Intel PrimeSense sensor.
Again, we use the best performing network according to Tab.~\ref{tab:quant_completion_archVariants}.
We see that our model, trained only on synthetic data, learns to generalize and transfer to real data.

\begin{figure*}[htb!]
	\vspace{-0.6cm}
	\centering
	\includegraphics[width=0.9\textwidth]{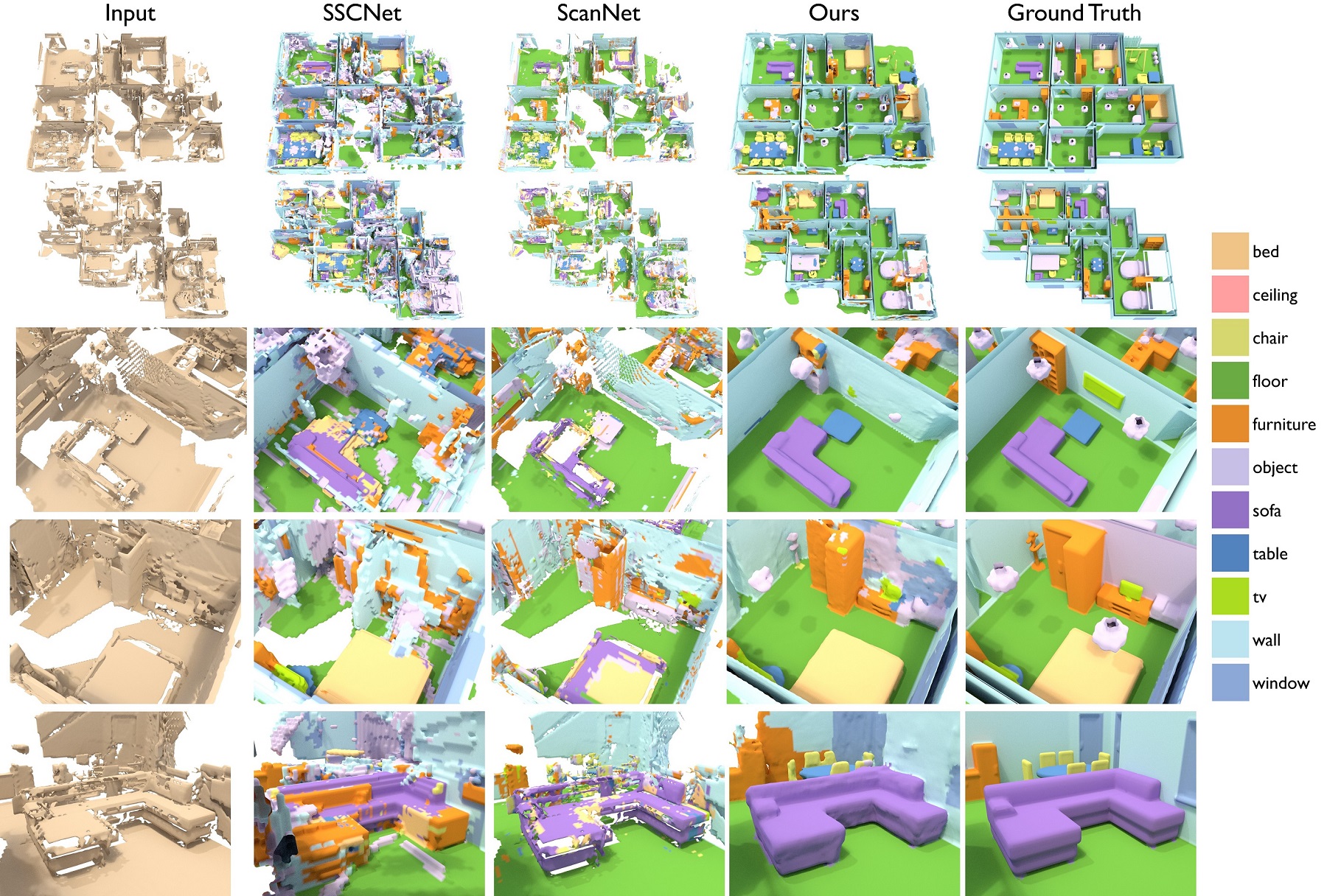}
	\caption{Semantic voxel labeling results on SUNCG; from left to right: input, SSCNet \cite{song2017ssc}, ScanNet \cite{dai2017scannet}, Ours, and ground truth.}
	\label{fig:results_semantics_synth}
\end{figure*}

\paragraph{Semantic Inference on SUNCG}

In Tab.~\ref{tab:quant_semantic}, we evaluate and compare our semantic segmentation on the SUNCG dataset. 
All methods were trained on the train set of scenes used by SSCNet~\cite{song2017ssc} and evaluated on the test set.
We use the SUNCG 11-label set.
Our semantic inference benefits significantly from the joint completion and semantic task, significantly outperforming current state of the art.

Fig.~\ref{fig:results_semantics_synth} shows qualitative semantic segmentation results on SUNCG scenes. Our ability to process the entire scene at test time, in contrast to previous methods which operate on fixed subvolumes, along with the autoregressive, joint completion task, produces more globally consistent and accurate voxel labels.

For semantic inference on real scans, we refer to the appendix. 

\section{Conclusion and Future Work}
\label{sec:conclusion}

In this paper, we have presented \OURS{}, a novel data-driven approach that takes an input partial 3D scan and predicts both completed geometry and semantic voxel labels for the entire scene at once.
The key idea is to use a fully-convolutional network that decouples train and test resolutions, thus allowing for variably-sized test scenes with unbounded spatial extents.
In addition, we use a coarse-to-fine prediction strategy combined with a volumetric autoregressive network that leverages large spatial contexts while simultaneously predicting local detail.
As a result, we achieve both unprecedented scene completion results as well as volumetric semantic segmentation with significantly higher accuracy than previous state of the art.

Our work is only a starting point for obtaining high-quality 3D scans from partial inputs, which is a typical problem for RGB-D reconstructions.
One important aspect for future work is to further improve output resolution.
Currently, our final output resolution of $\sim 5\mathrm{cm}^3$ voxels is still not enough---ideally, we would use even higher resolutions in order to resolve fine-scale objects, e.g., cups.
In addition, we believe that end-to-end training across all hierarchy levels would further improve performance with the right joint optimization strategy.
Nonetheless, we believe that we have set an important baseline for completing entire scenes.
We hope that the community further engages in this exciting task, and we are convinced that we will see many improvements along these directions.

\section*{Acknowledgments}
This work was supported by a Google Research Grant, a Stanford Graduate Fellowship, and a TUM-IAS Rudolf M{\"o}{\ss}bauer Fellowship.
We would also like to thank Shuran Song for helping with the SSCNet comparison.

{\small
\bibliographystyle{ieee}
\bibliography{main}
}

\clearpage
\clearpage
\begin{appendix}
In this appendix, we provide additional details for our \OURS{} submission.
First, we show a qualitative evaluation on real-world RGB-D data; see Sec.~\ref{sec:realqualitative}.
Second, we evaluate our semantics predictions on real-world benchmarks; see Sec.~\ref{sec:quantitative}.
Further, we provide details on the comparisons to Dai et al.~\cite{dai2017complete} in Sec.~\ref{sec:epn} and visualize the subvolume blocks used for the training of our spatially-invariant network in Sec.~\ref{sec:trainingblockpairs}.
In Sec.~\ref{sec:timings}, we compare the timings of our network against previous approaches showing that we not only outperform them in terms of accuracy and qualitative results, but also have a significant run-time advantage due to our architecture design.
Finally, we show additional results on synthetic data for completion and semantics in Sec.~\ref{sec:suncg-additional}.

\section{Qualitative Evaluation Real Data}
\label{sec:realqualitative}
\begin{figure*}[htb!] 
	\vspace{-0.5cm}
	\begin{centering}
		\includegraphics[width=0.98\linewidth]{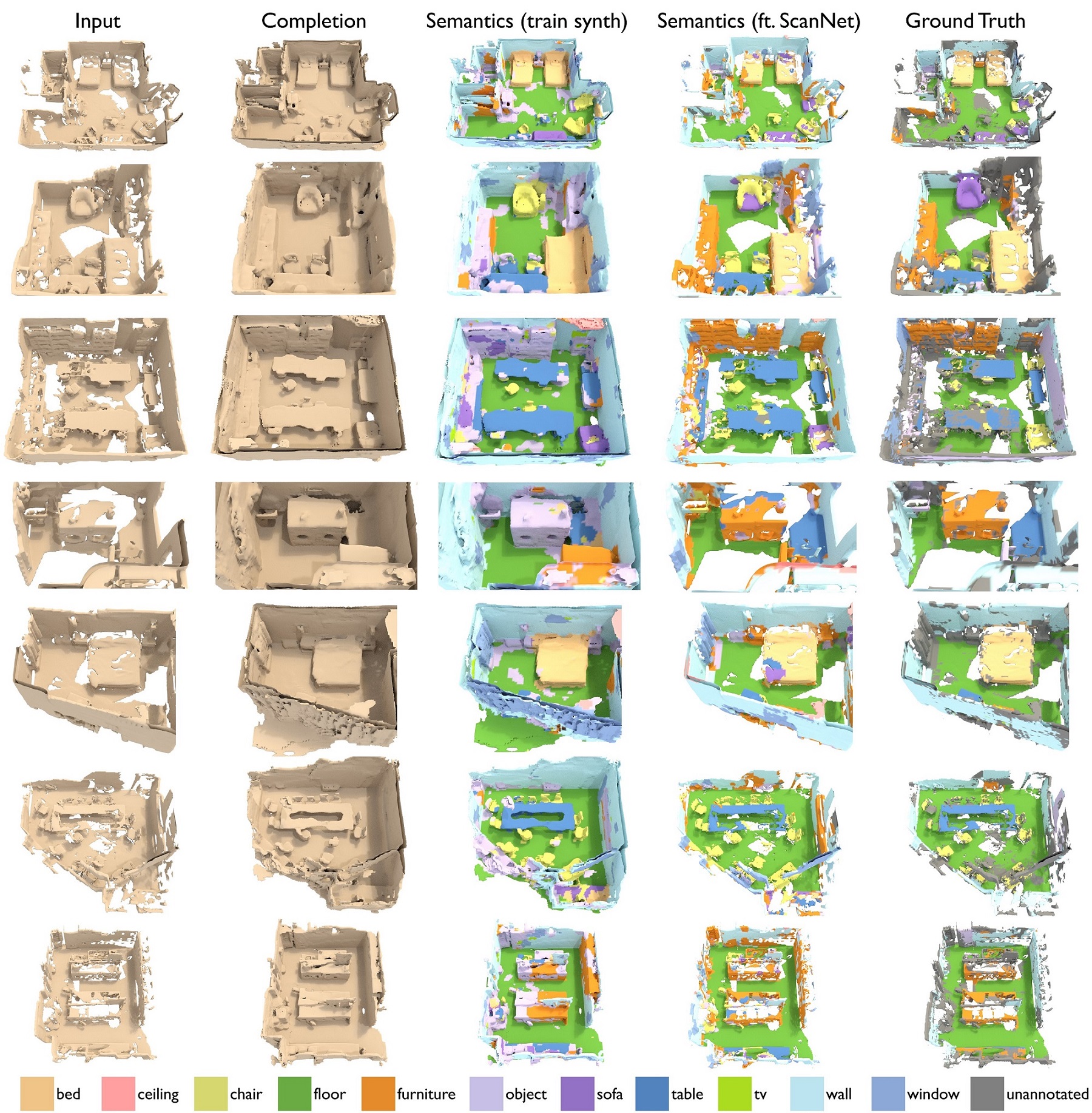}
		\caption{Additional results on ScanNet for our completion and semantic voxel labeling predictions.}
		\label{fig:scannet-additional}
		\vspace{-0.2cm}
	\end{centering}
\end{figure*}

\begin{figure*}[htb!] 
	\vspace{-0.5cm}
	\begin{centering}
		\includegraphics[width=0.8\linewidth]{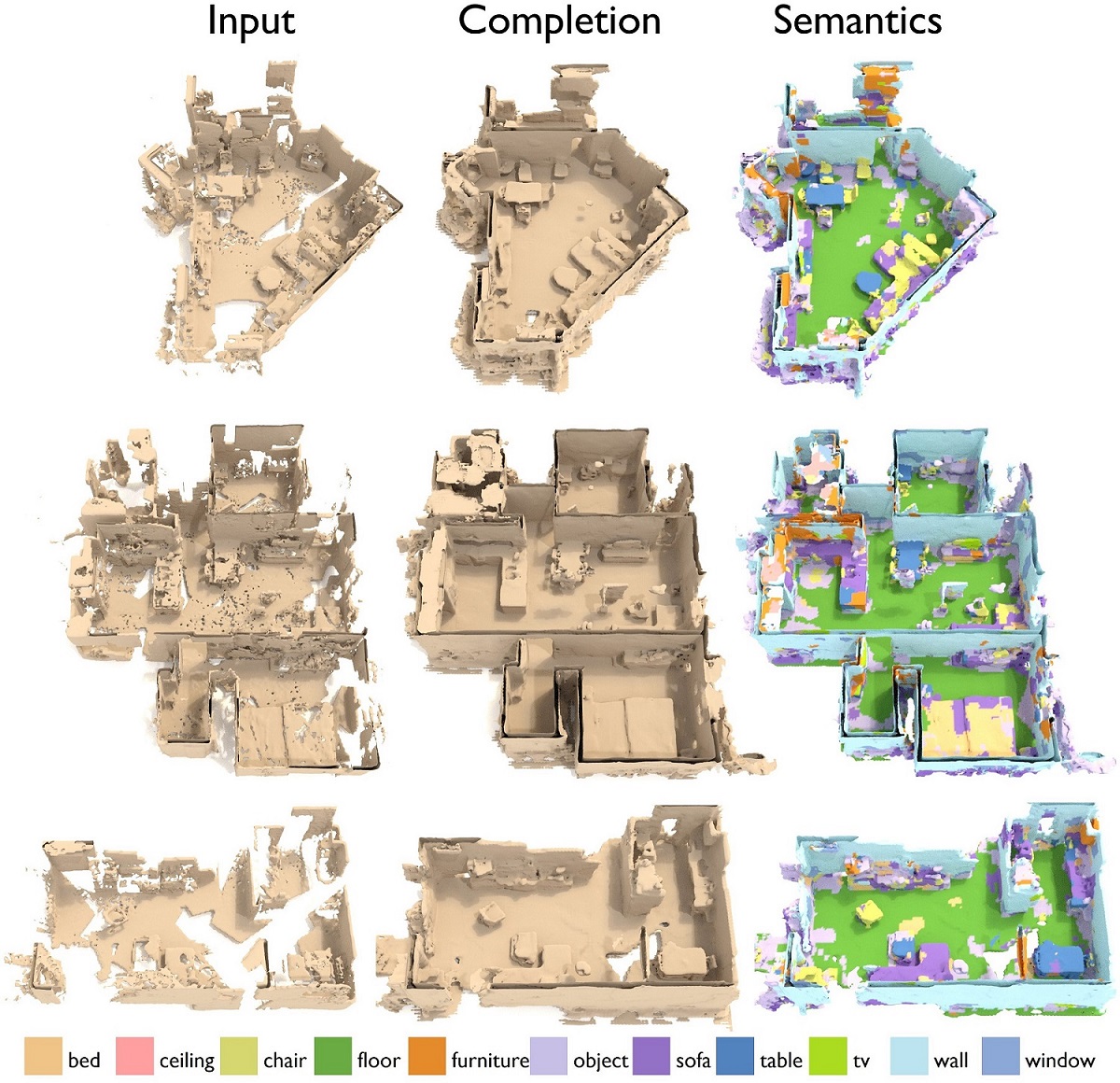}
		\caption{Additional results on Google Tango scans for our completion and semantic voxel labeling predictions.}
		\label{fig:tango-additional}
		\vspace{-0.2cm}
	\end{centering}
\end{figure*}

In Fig.~\ref{fig:scannet-additional} and Fig.~\ref{fig:tango-additional}, we use our network which is trained only on the synthetic SUNCG set, and use it infer missing geometry in real-world RGB-D scans; in addition, we infer per-voxel semantics.
We show results on several scenes on the publicly-available ScanNet~\cite{dai2017scannet} dataset;
the figure visualizes real input, completion (synthetically-trained), semantics (synthetically-trained), and semantics (synthetically pre-trained and fine-tuned on the ScanNet annotations).

\section{Quantitative Evaluation on Real Data}
\label{sec:quantitative}

For evaluation of semantic predictions on real-world scans, we provide a comprehensive comparison on the ScanNet \cite{dai2017scannet} and Matterport3D \cite{Matterport3D} datasets, which both have ground truth per-voxel annotations.
The results are shown in Tab.~\ref{tab:quant_semantic_real}.
We show results for our approach that is only trained on the synthetic SUNCG data; in addition, we fine-tune our semantics-only network on the respective real data.
Unfortunately, fine-tuning on real data is challenging when using a distance field representation given that the ground truth data is incomplete.
However, we can use pseudo-ground truth when leaving out frames and corresponding it to a more (but still not entirely) complete reconstruction when using an occupancy grid representation.
This strategy works on the Matterport3D dataset, as we have relatively complete scans to begin with; however, it is not applicably to the more incomplete ScanNet data.

\begin{table*}[htb!]
	\begin{center}
		\small
		\begin{tabular}{| c | c | c | c | c | c | c | c | c | c | c | c || c |}
			\hline
			\multicolumn{13}{|c|}{ScanNet}\\
			\hline
			& bed & ceil. & chair & floor & furn. & obj. & sofa & table & tv & wall & wind. & avg\\ \hline
			ScanNet~\cite{dai2017scannet} & {\bf 60.6} & 47.7 & {\bf 76.9} & {\bf 90.8} & {\bf 61.6} & {\bf 28.2} & 75.8 & 67.7 & 6.3 & 81.9 & 25.1 & 56.6\\ \hline
			Ours (SUNCG) & 42.6 & 69.5 & 53.1 & 70.9 & 23.7 & 20.0 & {\bf 76.3} & 63.4 & {\bf 29.1} & 57.0 & 26.9 & 48.4\\ \hline
			Ours (ft. ScanNet; sem-only) & 52.8 & {\bf 85.4} & 60.3 & 90.2 & 51.6 & 15.7 & 72.5 & {\bf 71.4} & 21.3 & {\bf 88.8} & {\bf 36.1} & {\bf 58.7}\\ \hline
			\hline
			\multicolumn{13}{|c|}{Matterport3D}\\
			\hline
			& bed & ceil. & chair & floor & furn. & obj. & sofa & table & tv & wall & wind. & avg\\ \hline
			Matterport3D~\cite{Matterport3D} & {\bf 62.8} & 0.1 & 20.2 & 92.4 & {\bf 64.3} & 17.0 & 27.7 & 10.7 & {\bf 5.5} & 76.4 & 15.0 & 35.7\\ \hline
			Ours (Matterport3D; sem-only) & 38.4 & 93.2 & {\bf 62.4} & 94.2 & 33.6 & {\bf 54.6} & 15.6 & {\bf 40.2} & 0.7 & 51.8 & {\bf 38.0} & 47.5\\ \hline
			Ours (Matterport3D) & 41.8 & {\bf 93.5} & 58.0 & {\bf 95.8} & 38.3 & 31.6 & {\bf 33.1} & 37.1 & 0.01 & {\bf 84.5} & 17.7 & {\bf 48.3}\\ \hline
		\end{tabular}
	\caption{Semantic labeling accuracy on real-world RGB-D. Per-voxel class accuracies on Matterport3D~\cite{Matterport3D} and ScanNet~\cite{dai2017scannet} test scenes. We can see a significant improvement on the average class accuracy on the Matterport3D dataset. }
		\label{tab:quant_semantic_real}	
		\vspace{-0.5cm}	
	\end{center}
\end{table*}

\section{Comparison Encoder-Predictor Network}
\label{sec:epn}

\begin{figure}[htb!] 
	\begin{centering}
		\includegraphics[width=\linewidth]{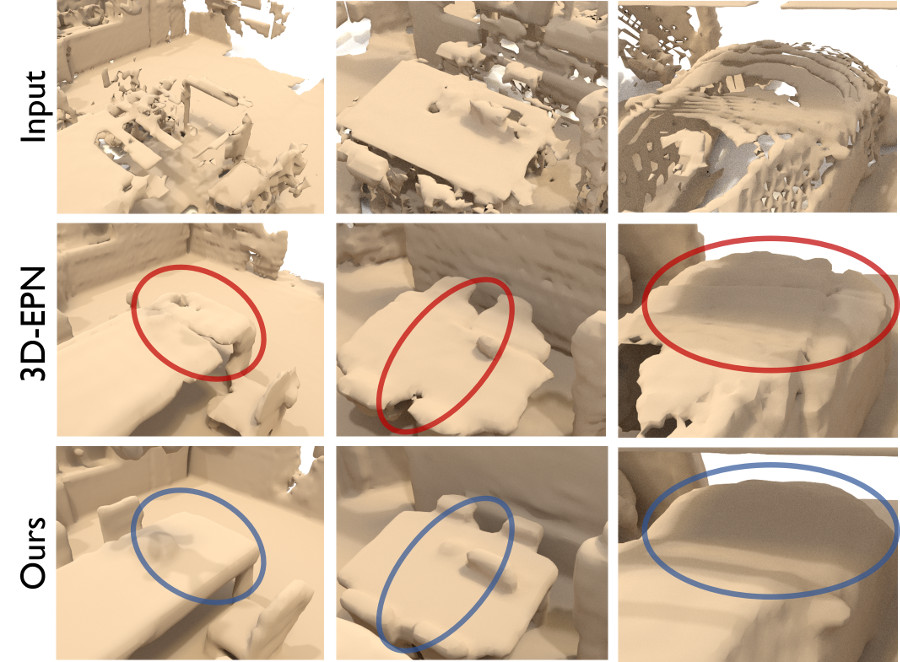}
		\caption{Applying the 3D-EPN approach~\cite{dai2017complete} to a scene by iteratively, independently predicting fixed-size subvolumes results in seams due to inconsistent predictions. Our approach, taking the entire partial scan as input, effectively alleviates these artifacts.}
		\label{fig:seams}
		\vspace{-0.2cm}
	\end{centering}
\end{figure}

In Fig.~\ref{fig:seams}, we visualize the problems of existing completion approach by Dai et al.~\cite{dai2017complete}.
They propose a 3D encoder-predictor network (3D-EPN), which takes as input a partial scan of an object and predicts the completed counterpart.
Their main disadvantage is that block predictions operate independently; hence, they do not consider information of neighboring blocks, which causes seams on the block boundaries.
Even though the quantitative error metrics are not too bad for the baseline approach, the visual inspection reveals that the boundary artifacts introduced at these seams are problematic.

\section{Training Block Pairs}
\label{sec:trainingblockpairs}

\begin{figure}[t] 
	\includegraphics[width=\linewidth]{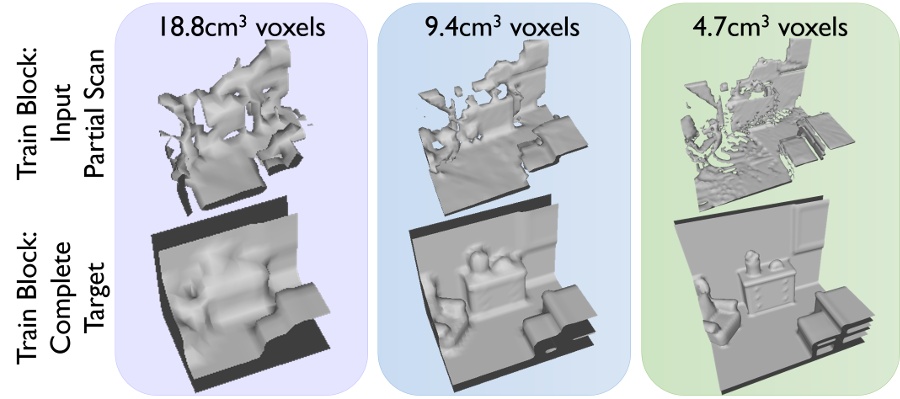}
	\caption{Subvolume train-test pairs of our three hierarchy levels.}
	\label{fig:train_blocks}
	\vspace{-0.2cm}
\end{figure}

In Fig.~\ref{fig:train_blocks}, we visualize the subvolumes used for training our fully-convolutional network on the three hierarchy levels of our network.
By randomly selecting a large variety of these subvolumes as ground truth pairs for training, we are able train our network such that it generalizes to varying spatial extents at test time.
Note again the fully-convolutional nature of our architecture, which allow the precessing of arbitrarily-sized 3D environments in a single test pass.

\section{Timings}
\label{sec:timings}

We evaluate the run-time performance of our method in Tab.~\ref{tab:timing} using an Nvidia GTX 1080 GPU.
We compare against the baseline 3D-EPN completion approach \cite{dai2017complete}, as well as the ScanNet semantic voxel prediction method \cite{dai2017scannet}.
The advantage of our approach is that our fully-convolutional architecture can process and entire scene at once.
Since we are using three hierarchy levels and an auto-regressive model with eight voxel groups, our method requires to run a total of $3 \times 8$ forward passes; however, note again that each of these passes is run over entire scenes.
In comparison, the ScanNet voxel labeling method is run on a per-voxel column basis.
That is, the $x-y$-resolution of the voxel grid determines the number of forward passes, which makes its runtime significantly slower than our approach even though the network architecture is less powerful (e.g., it cannot address completion in the first place).

The original 3D-EPN completion method \cite{dai2017complete} operates on a $32^3$ voxel grid to predict the completion of a single model.
We adapted this approach in to run on full scenes; for efficiency reasons we change the voxel resolution to $32 \times 32 \times 64$ to cover the full height in a single pass.
This modified version is run on each block independently, and requires the same number of forward passes than voxel blocks.
In theory, the total could be similar to one pass on a single hierarchy level; however, the separation of forward passes across several smaller kernel calls -- rather than fewer big ones -- is significantly less efficient on GPUs (in particular on current deep learning frameworks).

\begin{table*}[htb!]
	\begin{center}
		\small
		\begin{tabular}{| c || c | c | c | c | c | c |}
			\hline
			& \#Convs & \multicolumn{4}{|c|}{Scene Size (voxels)} \\
			&  & $82\times 64\times 64$ & $100\times 64\times 114$ & $162\times 64\times 164$ & $204\times 64\times 222$\\ \hline
			3D-EPN~\cite{dai2017complete} & 8 $+$ 2fc & 20.4 & 40.4 & 79.6 & 100.5 \\ \hline
			ScanNet~\cite{dai2017scannet} & 9 $+$ 2fc & 5.9 & 19.8 & 32.5 & 67.2 \\ \hline
			Ours (base level) & 32 & 0.4 & 0.4 & 0.6 & 0.9 \\ 
			Ours (mid level) & 42 & 0.7 & 1.3 & 2.2 & 4.7 \\ 
			Ours (high level) & 42 & 3.1 & 7.8 & 14.8 & 31.6 \\ \hline
			Ours (total) & - & 4.2 & 9.5 & 17.6 & 37.3 \\ \hline
		\end{tabular}
		\caption{Time (seconds) to evaluate test scenes of various sizes measured on a GTX 1080.}
		\label{tab:timing}	
		\vspace{-0.5cm}	
	\end{center}
\end{table*}

\section{Additional Results on Completion and Semantics on SUNCG}
\label{sec:suncg-additional}

\begin{figure*}[htb!] 
	\vspace{-0.5cm}
	\begin{centering}
		\includegraphics[width=0.98\linewidth]{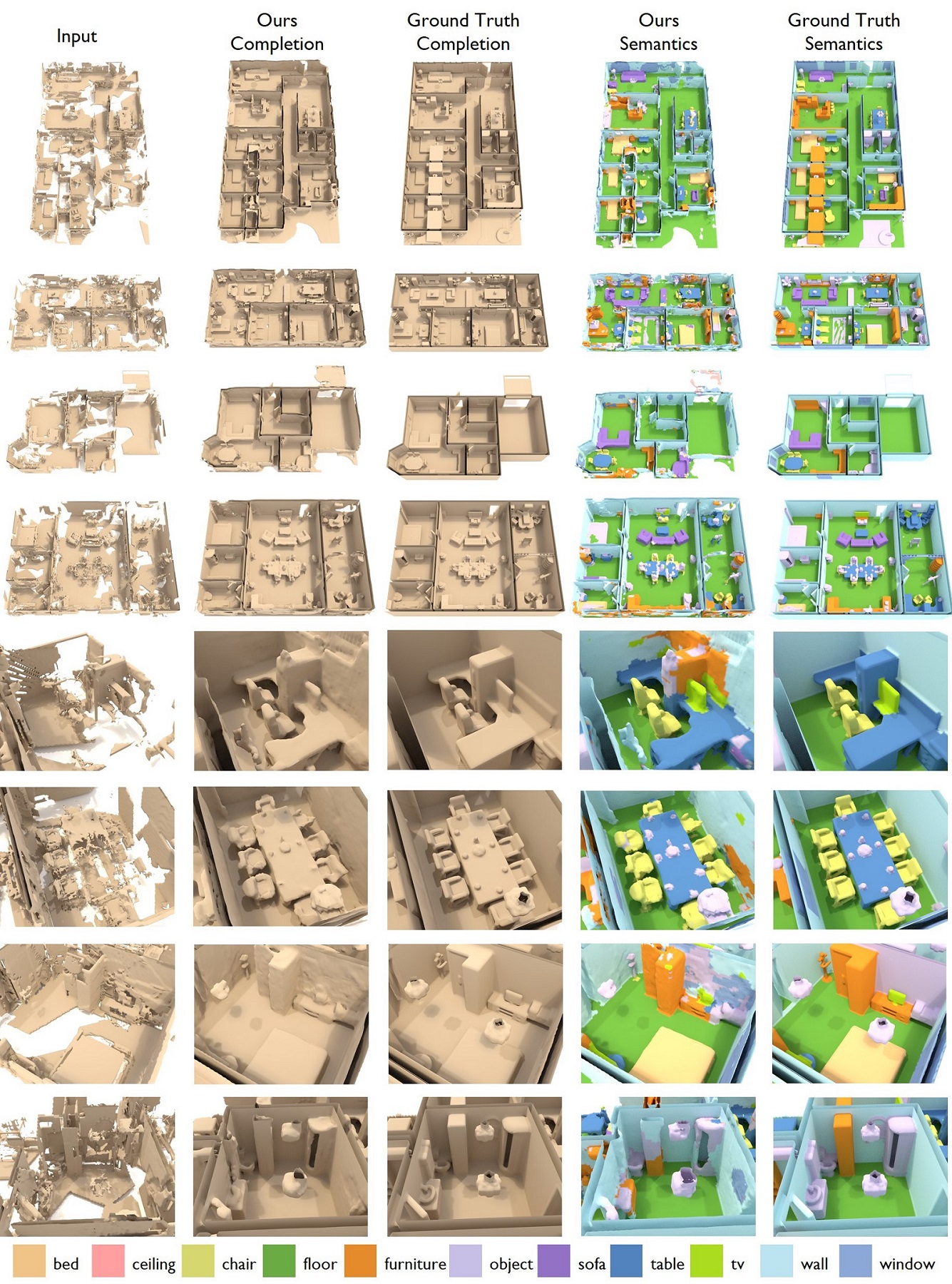}
		\caption{Additional results on SUNCG for our completion and semantic voxel labeling predictions.}
		\label{fig:suncg-additional}
		\vspace{-0.2cm}
	\end{centering}
\end{figure*}

Fig.~\ref{fig:suncg-additional} shows additional qualitative results for both completion and semantic predictions on the SUNCG dataset  \cite{song2017ssc}.
We show entire scenes as well as close ups spanning a variety of challenging scenarios.

\begin{table*}[htb!]
	\begin{center}
		\small
		\begin{tabular}{| c | c | c | c | c | c | c | c | c | c | c | c || c |}
			\hline
			& bed & ceil. & chair & floor & furn. & obj. & sofa & table & tv & wall & wind. & avg\\ \hline
			ScanNet~\cite{dai2017scannet} & 11.7 & 88.7 & 13.2 & 81.3 & 11.8 & 13.4 & 25.2 & 18.7 & 4.2 & 53.5 & 0.5 & 29.3\\ \hline
			SSCNet~\cite{song2017ssc} & 33.1 & 42.4 & 21.4 & 42.0 & 24.7 & 8.6 & 39.3 & 25.2 & 13.3 & 47.7 & 24.1 & 29.3\\ \hline
			{\bf Ours} & {\bf 50.4} & {\bf 95.5} & {\bf 35.3} & {\bf 89.4} & {\bf 45.2} & {\bf 31.3} & {\bf 57.4} & {\bf 38.2} & {\bf 16.7} & {\bf 72.2} & {\bf 33.3} & {\bf 51.4}\\ \hline
		\end{tabular}
		\caption{Semantic labeling on SUNCG scenes, measured as IOU per class over the visible surface of the partial test scans.}
		\label{tab:quant_semantic_iou}	
		\vspace{-0.5cm}	
	\end{center}
\end{table*}

\end{appendix}

\end{document}